\documentclass{article}
\usepackage{spconf,amsmath,graphicx,algorithm,algpseudocode}
\usepackage{float}
\usepackage{tikz}
\usepackage{pgfplots}
\usetikzlibrary{pgfplots.groupplots}
\usepackage[font=small,skip=0pt]{caption}
\pdfoutput=1

\title{Discriminative Acoustic Word Embeddings:  \\ Recurrent Neural Network-Based Approaches}

\name{Shane Settle, Karen Livescu\thanks{This research was supported by a Google faculty research award and NSF grant IIS-1321015. The opinions expressed in this work are those of the authors and do not necessarily reflect the views of the funding agency.}}
\address{Toyota Technological Institute at Chicago\\
 {\small \tt \{settle.shane, klivescu\}@ttic.edu}}

\begin{document}

\maketitle

\begin{abstract}
Acoustic word embeddings --- fixed-dimensional vector representations of variable-length spoken word segments --- have begun to be considered for tasks such as speech recognition and query-by-example search.  Such embeddings can be learned discriminatively so that they are similar for speech segments corresponding to the same word, while being dissimilar for segments corresponding to different words.  Recent work has found that acoustic word embeddings can outperform dynamic time warping on query-by-example search and related word discrimination tasks.  However, the space of embedding models and training approaches is still relatively unexplored.  In this paper we present new discriminative embedding models based on recurrent neural networks (RNNs).  We consider training losses that have been successful in prior work, in particular a cross entropy loss for word classification and a contrastive loss that explicitly aims to separate same-word and different-word pairs in a "Siamese network" training setting.
We find that both classifier-based and Siamese RNN embeddings improve over previously reported results on a word discrimination task, with Siamese RNNs outperforming classification models.  In addition, we present analyses of the learned embeddings and the effects of variables such as dimensionality and network structure.  
\end{abstract}

\section{Introduction}
\label{sec:intro}

Many speech processing tasks -- such as automatic speech recognition or spoken term detection -- hinge on associating segments of speech signals with word labels.  In most systems developed for such tasks, words are broken down into sub-word units such as phones, and models are built for the individual units.  An alternative, which has been considered by some researchers, is to consider each entire word segment as a single unit, without assigning parts of it to sub-word units.
One motivation for the use of whole-word approaches is that they avoid the need for sub-word models.  This is helpful since, despite decades of work on sub-word modeling~\cite{PMLA,ostendorf:asru99}, it still poses significant challenges.  For example, speech processing systems are still hampered by differences in conversational pronunciations~\cite{livescu2012subword}.  A second motivation is that considering whole words at once allows us to consider a more flexible set of features and reason over longer time spans. \\
\indent Whole-word approaches typically involve, at some level, template matching.  For example, in template-based speech recognition~\cite{dewachter+etal_taslp07,heigold2012investigations}, word scores are computed from dynamic time warping (DTW) distances between an observed segment and training segments of the hypothesized word.  In query-by-example search, putative matches are typically found by measuring the DTW distance between the query and segments of the search database~\cite{metze+etal_icassp13,anguera_icassp12,zhang2012fast,szoke+etal_icassp15}.  In other words, whole-word approaches often boil down to making decisions about whether two segments are examples of the same word or not. \\
\indent An alternative to DTW that has begun to be explored is the use of acoustic word embeddings (AWEs), or vector representations of spoken word segments.  AWEs are representations that can be learned from data, ideally such that the embeddings of two segments corresponding to the same word are close, while embeddings of segments corresponding to different words are far apart.  Once word segments are represented via fixed-dimensional embeddings, computing distances is as simple as measuring a cosine or Euclidean distance between two vectors.\\
\indent There has been some, thus far limited, work on acoustic word embeddings, focused on a number of embedding models, training approaches, and tasks~\cite{maas+etal_icmlwrl12,bengio+heigold_interspeech14,levin+etal_asru13,levin+etal_icassp15,kamper2016deep,chung2016unsupervised,voinea2014word,guoguo+etal_icassp15}.  In this paper we explore new embedding models based on recurrent neural networks (RNNs), applied to a word discrimination task related to query-by-example search.  RNNs are a natural model class for acoustic word embeddings, since they can handle arbitrary-length sequences.  We compare several types of RNN-based embeddings and analyze their properties.  Compared to prior embeddings tested on the same task, our best models achieve sizable improvements in average precision.

\section{Related work}
\label{sec:related}

We next briefly describe the most closely related prior work.

Maas {\it et al.}~\cite{maas+etal_icmlwrl12} and Bengio and Heigold~\cite{bengio+heigold_interspeech14} used acoustic word embeddings, based on convolutional neural networks (CNNs), to generate scores for word segments in automatic speech recognition.  Maas {\it et al.} trained CNNs to predict (continuous-valued) embeddings of the word labels, and used the resulting embeddings to define feature functions in a segmental conditional random field~\cite{zweig2009segmental} rescoring system.  Bengio and Heigold also developed CNN-based embeddings for lattice rescoring, but with a contrastive loss to separate embeddings of a given word from embeddings of other words.

Levin {\it et al.}~\cite{levin+etal_asru13} developed unsupervised embeddings based on representing each word as a vector of DTW distances to a collection of reference word segments.  This representation was subsequently used in several applications:  a segmental approach for query-by-example search~\cite{levin+etal_icassp15}, lexical clustering~\cite{kamper2014unsupervised}, and unsupervised speech recognition~\cite{kamper2015fully}.  Voinea {\it et al.}~\cite{voinea2014word} developed a representation also based on templates, in their case phone templates, designed to be invariant to specific transformations, and showed their robustness on digit classification.

Kamper {\it et al.}~\cite{kamper2016deep} compared several types of acoustic word embeddings for a word discrimination task related to query-by-example search, finding that embeddings based on convolutional neural networks (CNNs) trained with a contrastive loss outperformed the reference vector approach of Levin {\it et al.}~\cite{levin+etal_asru13} as well as several other CNN and DNN embeddings and DTW using several feature types.  There have now been a number of approaches compared on this same task and data~\cite{levin+etal_asru13,carlin+etal_icassp11,kamper+etal_icassp15,jansen+etal_icassp13b}.  For a direct comparison with this prior work, in this paper we use the same task and some of the same training losses as Kamper {\it et al.}, but develop new embedding models based on RNNs.

The only prior work of which we are aware using RNNs for acoustic word embeddings is that of Chen {\it et al.}~\cite{guoguo+etal_icassp15} and Chung {\it et al.}~\cite{chung2016unsupervised}.  Chen {\it et al.} learned a long short-term memory (LSTM) RNN for word classification and used the resulting hidden state vectors as a word embedding in a query-by-example task.  The setting was quite specific, however, with a small number of queries and speaker-dependent training.  Chung {\it et al.}~\cite{chung2016unsupervised} worked in an unsupervised setting and trained single-layer RNN autoencoders to produce embeddings for a word discrimination task.  In this paper we focus on the supervised setting, and compare a variety of RNN-based structures trained 
with different losses.

\vspace{-.1in}
\section{Approach}
\label{sec:models}
\vspace{-.1in}

\begin{figure}
  \centering
  \centerline{\includegraphics[width=9cm]{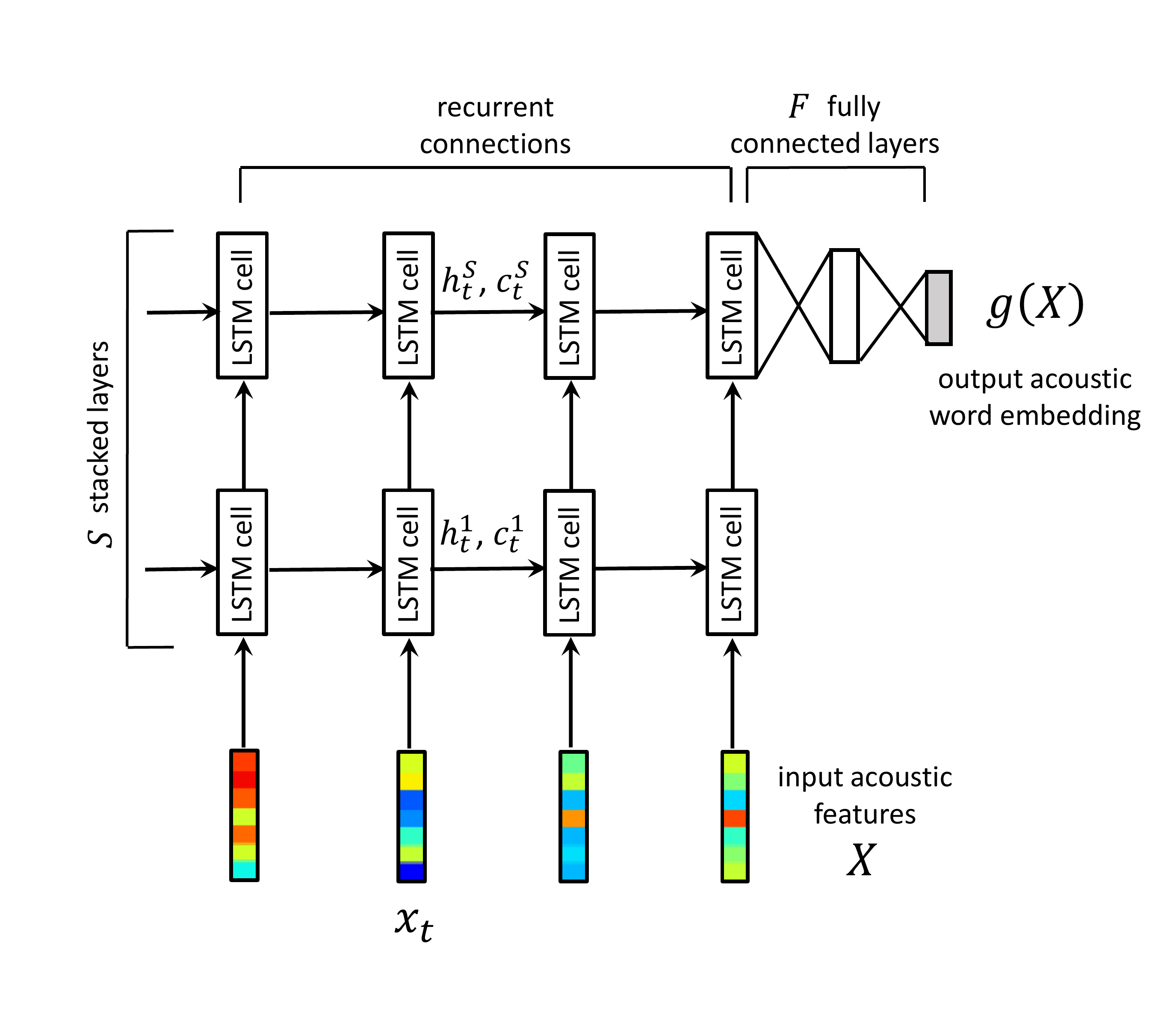}}\vspace{-.2in}
  \caption{LSTM-based acoustic word embedding model.  For GRU-based models, the structure is the same, but the LSTM cells are replaced with GRU cells, and there is no cell activation vector; the recurrent connections only carry the hidden state vector $\mathbf{h}_t^l$.}
  \label{fig:rnn}
\end{figure}

An acoustic word embedding is a function that takes as input a speech segment corresponding to a word, $X = \{x_t\}_{t=1}^T$, where each $x_t$ is a vector of frame-level acoustic features, and outputs a fixed-dimensional vector representing the segment, $g(X)$.  The basic embedding model structure we use is shown in Fig.~\ref{fig:rnn}.  The model consists of a deep RNN with some number $S$ of stacked layers, whose final hidden state vector is passed as input to a set of $F$ of fully connected layers; the output of the final fully connected layer is the embedding $g(X)$.

The RNN hidden state at each time frame can be viewed as a representation of the input seen thus far, and its value in the last time frame $T$ could itself serve as the final word embedding.  The fully connected layers are added to account for the fact that some additional transformation may improve the representation.  For example, the hidden state may need to be larger than the desired word embedding dimension, in order to be able to "remember" all of the needed intermediate information.  Some of that information may not be needed in the final embedding.  In addition, the information maintained in the hidden state may not necessarily be discriminative; some additional linear or non-linear transformation may help to learn a discriminative embedding.

Within this class of embedding models, we focus on Long Short-Term Memory (LSTM) networks~\cite{hochreiter1997long} and Gated Recurrent Unit (GRU) networks~\cite{chung2014empirical}.  These are both types of RNNs that include a mechanism for selectively retaining or discarding information at each time frame when updating the hidden state, in order to better utilize long-term context.  Both of these RNN variants have been used successfully in speech recognition~\cite{graves2013speech,sak2014long,chorowski2015attention,lu2015study}.

In an LSTM RNN, at each time frame both the hidden state $\mathbf{h}_t$ and an associated ``cell memory" vector $\mathbf{c}_t$, are updated and passed on to the next time frame.  In other words, each forward edge in Figure~\ref{fig:rnn} can be viewed as carrying both the cell memory and hidden state vectors.  The updates are modulated by the values of several gating vectors, which control the degree to which the cell memory and hidden state are updated in light of new information in the current frame.  For a single-layer LSTM network, the updates are as follows:
\vspace{-0.1cm}
       \begin{align*}
       	\mathbf{i}_t &=&& \sigma( \mathbf{W_i} [\mathbf{x}_t, \mathbf{h}_{t-1}] + \mathbf{b_i}) & \;\;\; & \textrm{input gate}\\
        \mathbf{f}_t &=&& \sigma( \mathbf{W_f} [\mathbf{x}_t, \mathbf{h}_{t-1}] + \mathbf{b_f}) & \;\;\; & \textrm{forget gate}\\
        \mathbf{\widetilde{c}}_t &=&& \tanh( \mathbf{W_c} [\mathbf{x}_t, \mathbf{h}_{t-1}] + \mathbf{b_c}) & \;\;\; & \textrm{candidate cell memory}\\
        \mathbf{c}_t &=&& \mathbf{i}_t \odot \mathbf{\widetilde{c}}_t + \mathbf{f}_t \odot \mathbf{c}_{t-1} & \;\;\; & \textrm{cell memory} \\
      	\mathbf{o}_t &=&& \sigma( \mathbf{W_o} [\mathbf{x}_t, \mathbf{h}_{t-1}] + \mathbf{b_o}) & \;\;\; & \textrm{output gate} \\
      	\mathbf{h}_t &=&& \mathbf{o}_t \odot \tanh(\mathbf{c}_t) & \;\;\; & \textrm{hidden state}
       \end{align*}

\noindent where $\mathbf{h}_t, \mathbf{c}_t, \mathbf{\widetilde{c}}_t, \mathbf{i}_t, \mathbf{f}_t$, and $\mathbf{o}_t$ are all vectors of the same dimensionality, $\mathbf{W_i}, \mathbf{W_o}, \mathbf{W_f}$, and $\mathbf{W_c}$ are learned weight matrices of the appropriate sizes, $\mathbf{b_i}, \mathbf{b_o}, \mathbf{b_f}$ and $\mathbf{b_c}$ are learned bias vectors, $\sigma(\cdot)$ is a componentwise logistic activation, and $\odot$ refers to the Hadamard (componentwise) product.  
    
Similarly, in a GRU network, at each time step a GRU cell determines what components of old information are retained, overwritten, or modified in light of the next step in the input sequence.  The output from a GRU cell is only the hidden state vector.  A GRU cell uses a reset gate $\mathbf{r}_t$ and an update gate $\mathbf{u}_t$ as described below for a single-layer network:
	\begin{align*}
    	\mathbf{r}_t &=&& \sigma( \mathbf{W_r} [\mathbf{x}_t, \mathbf{h}_{t-1}] + \mathbf{b_r}) & \;\;\; & \textrm{reset gate}\\
    	\mathbf{u}_t &=&& \sigma( \mathbf{W_u} [\mathbf{x}_t, \mathbf{h}_{t-1}] + \mathbf{b_u}) & \;\;\; & \textrm{update gate}\\
    	\mathbf{\widetilde{h}}_t &=&& \tanh( \mathbf{W_h} [\mathbf{x}_t, \mathbf{r}_t \odot \mathbf{h}_{t-1}] + \mathbf{b_h}) & \;\;\; & \textrm{candidate hidden}\\
    	\mathbf{h}_t &=&& \mathbf{u}_t \odot \mathbf{h}_{t-1} + (1 - \mathbf{u}_t) \odot \mathbf{\widetilde{h}}_t & \;\;\; & \textrm{hidden state}
      \end{align*}
\noindent where $\mathbf{r}_t, \mathbf{u}_t, \mathbf{\widetilde{h}}_t$, and $\mathbf{h}_t$ are all the same dimensionality, $\mathbf{W_r}, \mathbf{W_u}$, and $\mathbf{W_h}$ are learned weight matrices of the appropriate size, and $\mathbf{b_r}$, $\mathbf{b_u}$ and $\mathbf{b_h}$ are learned bias vectors.

All of the above equations refer to single-layer networks.  In a deep network, with multiple stacked layers, the same update equations are used in each layer, with the state, cell, and gate vectors replaced by layer-specific vectors $\mathbf{h}_t^l, \mathbf{c}_t^l,$ and so on for layer $l$.  For all but the first layer, the input $\mathbf{x}_t$ is replaced by the hidden state vector from the previous layer $\mathbf{h}_t^{l-1}$.

For the fully connected layers, we use rectified linear unit (ReLU)~\cite{nair2010rectified} activation, except for the final layer which depends on the form of supervision and loss used in training.

\vspace{-.1in}
\subsection{Training}
\label{approach-training}
We train the RNN-based embedding models using a set of pre-segmented spoken words.  We use two main training approaches, inspired by prior work but with some differences in the details.  As in~\cite{kamper2016deep,bengio+heigold_interspeech14}, our first approach is to use the word labels of the training segments and train the networks to classify the word.  In this case, the final layer of $g(X)$ is a log-softmax layer.  Here we are limited to the subset of the training set that has a sufficient number of segments per word to train a good classifier, and the output dimensionality is equal to the number of words (but see~\cite{kamper2016deep} for a study of varying the dimensionality in such a classifier-based embedding model by introducing a bottleneck layer).  This model is trained end-to-end and is optimized with a cross entropy loss.  Although labeled data is necessarily limited, the hope is that the learned models will be useful even when applied to spoken examples of words not previously seen in the training data.  For words not seen in training, the embeddings should correspond to some measure of similarity of the word to the training words, measured via the posterior probabilities of the previously seen words.  In the experiments below, we examine this assumption by analyzing performance on words that appear in the training data compared to those that do not.

The second training approach, based on earlier work of Kamper {\it et al.}~\cite{kamper2016deep}, is to train "Siamese" networks~\cite{bromley+etal_ijpr93}.  In this approach, full supervision is not needed; rather, we use weak supervision in the form of pairs of segments labeled as same or different.  The base model remains the same as before---an RNN followed by a set of fully connected layers---but the final layer is no longer a softmax but rather a linear activation layer of arbitrary size.  In order to learn the parameters, we simultaneously feed three word segments through three copies of our model (i.e.~three networks with shared weights).  One input segment is an ``anchor", $x_{a}$, the second is another segment with the same word label, $x_{s}$, and the third is a segment corresponding to a different word label, $x_{d}$.  Then, the network is trained using a ``cos-hinge" loss:
\vspace{-.05in}
\begin{equation*}
l_{\textrm{cos hinge}} = \max\{0, m + d_{cos}(x_a, x_s) - d_{cos}(x_a, x_d)\}
\label{eq:cos-hinge}
\end{equation*}
\noindent where $d_{cos}(x_1,x_2) = 1 - \cos(x_1,x_2)$ is the cosine distance between $x_1,x_2$.  Unlike cross entropy training, here we directly aim to optimize relative (cosine) distance between same and different word pairs.  For tasks such as query-by-example search, this training loss better respects our end objective, and can use more data since neither fully labeled data nor any minimum number of examples of each word should be needed.

\vspace{-.1in}
\section{EXPERIMENTS}
\label{sec:format}
\vspace{-.05in}

Our end goal is to improve performance on downstream tasks requiring accurate word discrimination.  In this paper we use an intermediate task that more directly tests whether same- and different-word pairs have the expected relationship. and that allows us to compare to a variety of prior work.  Specifically, we use the word discrimination task of Carlin {\it et al.}~\cite{carlin+etal_icassp11}, which is similar to a query-by-example task where the word segmentations are known.  The evaluation consists of determining, for each pair of evaluation segments, whether they are examples of the same or different words, and measuring performance via the average precision (AP).  We do this by measuring the cosine similarity between their acoustic word embeddings and declaring them to be the same if the distance is below a threshold.  By sweeping the threshold, we obtain a precision-recall curve from which we compute the AP.

The data used for this task is drawn from the Switchboard conversational English corpus~\cite{godfrey1992switchboard}.  The word segments range from 50 to 200 frames in length.  The acoustic features in each frame (the input to the word embedding models $x_t$) are 39-dimensional MFCCs+$\Delta$+$\Delta \Delta$.
We use the same train, development, and test partitions as in prior work \cite{kamper2016deep,levin+etal_asru13}, and the same acoustic features as in~\cite{kamper2016deep}, for as direct a comparison as possible.  The train set contains approximately 10k example segments, while dev and test each contain approximately 11k segments (corresponding to about 60M pairs for computing the dev/test AP).  As in~\cite{kamper2016deep}, when training the classification-based embeddings, we use a subset of the training set containing all word types with a minimum of 3 occurrences, reducing the training set size to approximately 9k segments.\footnote{We thank Herman Kamper for assistance with the data and evaluation.}

When training the Siamese networks, the training data consists of all of the same-word pairs in the full training set (approximately 100k pairs).  For each such training pair, we randomly sample a third example belonging to a different word type, as required for the $l_{\textrm{cos hinge}}$ loss. 

\vspace{-.1in}
\subsection{Classification network details}
\label{ssec:classification}

Our classifier-based embeddings use LSTM or GRU networks with 2--4 stacked layers and 1--3 fully connected layers.  The final embedding dimensionality is equal to the number of unique word labels in the training set, which is 1061.  The recurrent hidden state dimensionality is fixed at 512 and dropout~\cite{srivastava2014dropout} between stacked recurrent layers is used with probability $p = 0.3$.  The fully connected hidden layer dimensionality is fixed at 1024.  Rectified linear unit (ReLU) non-linearities and dropout with $p = 0.5$ are used between fully-connected layers.  However, between the final recurrent hidden state output and the first fully-connected layer no non-linearity or dropout is applied.  These settings were determined through experiments on the development set.

The classifier network is trained with a cross entropy loss and optimized using stochastic gradient descent (SGD) with Nesterov momentum \cite{nesterov1983method}.  The learning rate is initialized at 0.1 and is reduced by a factor of 10 according to the following heuristic:  If 99\% of the current epoch's average batch loss is greater than the running average of batch losses over the last 3 epochs, this is considered a plateau; if there are 3 consecutive plateau epochs, then the learning rate is reduced.  Training stops when reducing the learning rate no longer improves dev set AP.  Then, the model from the epoch corresponding to the the best dev set AP is chosen.  Several other optimizers---Adagrad \cite{Duchi:EECS-2010-24}, Adadelta \cite{DBLP:journals/corr/abs-1212-5701}, and Adam \cite{DBLP:journals/corr/KingmaB14}---were explored in initial experiments on the dev set, but all reported results were obtained using SGD with Nesterov momentum.  

\vspace{-.1in}
\subsection{Siamese network details}
\label{ssec:siamese}

For experiments with Siamese networks, we initialize (warm-start) the networks with the tuned classification network, removing the final log-softmax layer and replacing it with a linear layer of size equal to the desired embedding dimensionality.  We explored embeddings with dimensionalities between 8 and 2048.  We use a margin of 0.4 in the cos-hinge loss.

In training the Siamese networks, each training mini-batch consists of $2B$ triplets. $B$ triplets are of the form $(x_a,x_s,x_d)$ where $x_a$ and $x_s$ are examples of the same class (a pair from the 100k same-word pair set) and $x_d$ is a randomly sampled example from a different class. Then, for each of these $B$ triplets $(x_a,x_s,x_d)$, an additional triplet $(x_s,x_a,x_d)$ is added to the mini-batch to allow all segments to serve as anchors.  This is a slight departure from earlier work~\cite{kamper2016deep}, which we found to improve stability in training and performance on the development set.

In preliminary experiments, we compared two methods for choosing the negative examples $x_d$ during training, a uniform sampling approach and a non-uniform one. In the case of uniform sampling, we sample $x_d$ uniformly at random from the full set of training examples with labels different from $x_a$.  This sampling method requires only word-pair supervision.
In the case of non-uniform sampling, $x_d$ is sampled in two steps. First, we construct a distribution $P_{y|label(x_a)}$ over word labels $y$ and sample a different label from it.  Second, we sample an example uniformly from within the subset with the chosen label. The goal of this method is to speed up training by targeting pairs that violate the margin constraint.  To construct the multinomial PMF $P_{y|label(x_a)}$, we maintain an $n \times n$ matrix $\mathbf{S}$, where $n$ is the number of unique word labels in training. Each word label corresponds to an integer $i$ $\in$ [1, $n$] and therefore a row in $\mathbf{S}$.  The values in a row of $\mathbf{S}$ are considered similarity scores, and we can retrieve the desired PMF for each row by normalizing by its sum.

At the start of each epoch, we initialize $\mathbf{S}$ with $0$'s along the diagonal and $1$'s elsewhere (which reduces to uniform sampling).  For each training pair $(d_{cos}(x_a,x_s), d_{cos}(x_a,x_d))$, we update $\mathbf{S}$ for both $(i,j) = (label(x_a),label(x_d))$ and $(i,j) = (label(x_d),label(x_a))$:
\vspace{0in}
\[	s_{i,j} \mathrel{+}=  \begin{cases}
				cos(x_a,x_d) & d_{cos}(x_a,x_d) \leq d_{cos}(x_a,x_s) + m^* \\
                0 & \text{otherwise}
   			\end{cases}
\]
\noindent The PMFs $P_{y|label(x_a)}$ are updated after the forward pass of an entire mini-batch.  The constant $m^*$ enforces a potentially stronger constraint than is used in the $l_{\textrm{cos hinge}}$ loss, in order to promote diverse sampling. In all experiments, we set $m^* = 0.6$. This is a heuristic approach, and it would be interesting to consider various alternatives. Preliminary experiments showed that the non-uniform sampling method outperformed uniform sampling, and in the following we report results with non-uniform sampling. 

We optimize the Siamese network model using SGD with Nesterov momentum for 15 epochs. The learning rate is initialized to 0.001 and dropped every 3 epochs until no improvement is seen on the dev set.  The final model is taken from the epoch with the highest dev set AP.  All models were implemented in Torch \cite{collobert2011torch7} and used the rnn library of \cite{DBLP:journals/corr/LeonardWW15}.

\begin{table}[t]
\caption{\label{OverallAP} {Final test set results in terms of average precision (AP). Dimensionalities marked with * refer to dimensionality per frame for DTW-based approaches. For CNN and LSTM models, results are given as means over several training runs (5 and 10, respectively) along with their standard deviations.}}
  \vspace{2mm}
  \centerline{
\begin{tabular}{| l | r | l |}
  \hline
  \multicolumn{1}{|l|}{Model} &
  \multicolumn{1}{c|}{Dim} &
  \multicolumn{1}{c|}{AP} \\
  \hline \hline
  	MFCCs + DTW \cite{kamper2016deep} & $39^*$ & $0.214$ ~~~ \\
    Corr. autoencoder + DTW~\cite{kamper+etal_icassp15} & $100^*$ & $0.469$ ~~~ \\ \hline
    Classifier CNN \cite{kamper2016deep} & $1061$ & $0.532 \pm 0.014$ ~~~ \\
    Siamese CNN \cite{kamper2016deep} & $1024$ & $0.549 \pm 0.011$ ~~~ \\ \hline
    Classifier LSTM & $1061$ & $0.616 \pm 0.009$ ~~~ \\
    Siamese LSTM & $1024$ & $0.671 \pm 0.011$ ~~~ \\
  \hline
   \end{tabular}
 }
 \end{table}

\vspace{-.05in}
\section{Results}
\label{sec:results}
\vspace{-.05in}
Based on development set results, our final embedding models are LSTM networks with 3 stacked layers and 3 fully connected layers, with output dimensionality of 1024 in the case of Siamese networks.  Final test set results are given in Table~\ref{OverallAP}.  We include a comparison with the best prior results on this task from~\cite{kamper2016deep}, as well as the result of using standard DTW on the input MFCCs (reproduced from~\cite{kamper2016deep}) and the best prior result using DTW, obtained with frame features learned with correlated autoencoders~\cite{kamper+etal_icassp15}.  Both classifier and Siamese LSTM embedding models outperform all prior results on this task of which we are aware.\footnote{Yuan {\it et al.}~\cite{yuan2016learning} have recently been able to improve AP on this test set even further with CNN embeddings, by using a large set of additional (cross-lingual) training data.  We do not consider these results to be comparable because of their reliance on additional data.}

We next analyze the effects of model design choices, as well as the learned embeddings themselves.

\vspace{-.1in}
\subsection{Effect of model structure}
\label{ssec:structure}

Table~\ref{ClassifierAP} shows the effect on development set performance of the number of stacked layers $S$, the number of fully connected layers $F$, and LSTM vs.~GRU cells, for classifier-based embeddings.
The best performance in this experiment is achieved by the LSTM network with $S = F = 3$.  However, performance still seems to be improving with additional layers, suggesting that we may be able to further improve performance by adding even more layers of either type.  However, we fixed the model to $S = F = 3$ in order to allow for more experimentation and analysis within a reasonable time.

      \begin{table}[th]
        \caption{\label{ClassifierAP} {Average precision on the dev set, using classifier-based embeddings. $S$ = \# stacked layers, $F$ = \# fully connected layers.}}
        \vspace{2mm}
        \centerline{
          \begin{tabular}{| r | r | r | r |}
            \hline
            \multicolumn{1}{|c|}{$S$} &
            \multicolumn{1}{c|}{$F$} &
            \multicolumn{1}{c|}{GRU AP} &
            \multicolumn{1}{c|}{LSTM AP} \\
            \hline \hline
               	$2$ & $1$ & $0.213$ & $0.240$ ~~~ \\
            	$3$ & $1$ & $0.252$ & $0.244$ ~~~ \\
            	$4$ & $1$ & $0.303$ & $0.267$ ~~~ \\
                $3$ & $2$ & $0.412$ & $0.418$ ~~~ \\
            	$3$ & $3$ & $0.445$ & $0.519$ ~~~ \\
            \hline
          \end{tabular}
        }
      \end{table}

\begin{figure}
\centering
\begin{tikzpicture}
\begin{groupplot}[
    group style={
        group name= myplots,
        group size=2 by 1,
        horizontal sep=0pt,
        ylabels at= edge left,
        yticklabels at= edge left
    },
    width=5.3cm,
    height=5.3cm,
    ylabel=dev AP,
    ymin=0.4,
    ymax=1.0
]
\nextgroupplot[
    xmode= log,
    log basis x=2,
    xlabel= dim,
    legend style={at={(0,1)},anchor=north west},
    xmin=8,
    xmax=2048,
]
\addplot[mark=o, blue] coordinates {(8,0.3940) (16,0.5037) (32,0.5257) (64,0.5371)(128,0.546) (256,0.552) (512,0.557) (1024, 0.567) (2048,0.548)};
\addlegendentry{Siamese}
\addplot[mark=none, red] coordinates {(8,0.519) (16,0.519) (32,0.519) (64,0.519)(128,0.519) (256,0.519) (512,0.519) (1024, 0.519) (2048,0.519)};
\addlegendentry{Classifier}
\nextgroupplot[
xlabel= min. occurrence count,
legend style = {at={(0,1)},anchor=north west}
]
\addplot [blue,mark=o] coordinates {(0,0.557) (1,0.6944) (3,0.7225) (5,0.7700) (7,0.801) (10,0.833) (15,0.856)};
\addplot [red,mark=o] coordinates {(0,0.519) (1,0.603) (3,0.619) (5,0.648) (7,0.667) (10, 0.695) (15,0.717)};
\end{groupplot}
\end{tikzpicture}
\vspace{-.05in}
\caption{Effect of embedding dimensionality (left) and occurrences in training set (right).}
\label{fig:varydim_varymin}
\end{figure}
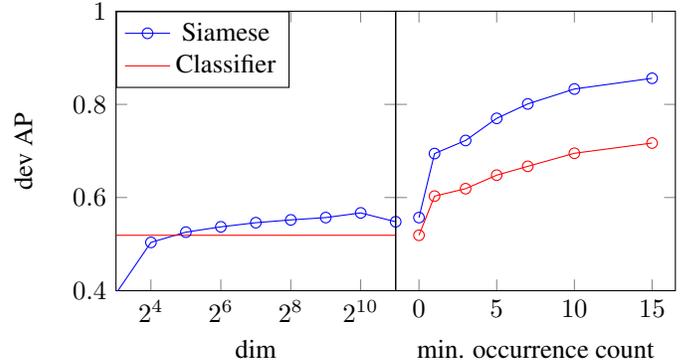

\begin{figure*}
\centering
\includegraphics[height=8cm,width=15cm]{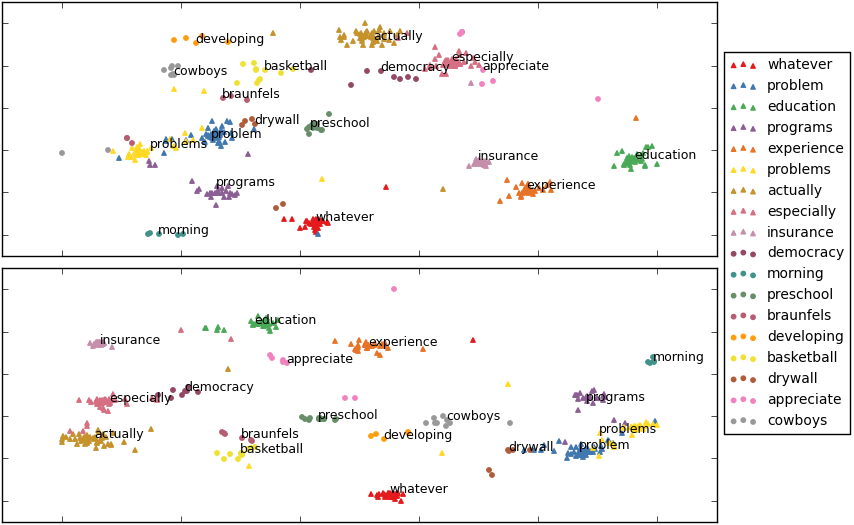}
\vspace{.05in}
\caption{t-SNE visualization of word embeddings from the dev set produced by the classifier (top) vs.~Siamese (bottom) models.  Word labels seen at training time are denoted by triangles and word labels unseen at training time are denoted by circles.}
\label{fig:tsne}
\end{figure*}

Table~\ref{ClassifierAP} reveals an interesting trend.  When only one fully connected layer is used, the GRU networks outperform the LSTMs given a sufficient number of stacked layers.  On the other hand, once we add more fully connected layers, the LSTMs outperform the GRUs.
In the first few lines of Table~\ref{ClassifierAP}, we use 2, 3, and 4 layer stacks of LSTMs and GRUs while holding fixed the number of fully-connected layers at $F = 1$. There is clear utility in stacking additional layers; however, even with 4 stacked layers the RNNs still underperform the CNN-based embeddings of~\cite{kamper2016deep} until we begin adding fully connected layers.

After exploring a variety of stacked RNNs, we fixed the stack to 3 layers and varied the number of fully connected layers. The value of each additional fully connected layer is clearly greater than that of adding stacked layers.  All networks trained with 2 or 3 fully connected layers obtain more than 0.4 AP on the development set, while stacked RNNs with 1 fully connected layer are at around 0.3 AP or less.  This may raise the question of whether some simple fully connected model may be all that is needed; however, previous work has shown that this approach is not competitive~\cite{kamper2016deep}, and convolutional or recurrent layers are needed to summarize arbitrary-length segments into a fixed-dimensional representation.

\vspace{-.1in}
\subsection{Effect of embedding dimensionality}
\label{ssec:varymincount}

For the Siamese networks, we varied the output embedding dimensionality, as shown in Fig.~\ref{fig:varydim_varymin}. This analysis shows that the embeddings learned by the Siamese RNN network are quite robust to reduced dimensionality, outperforming the classifier model for all dimensionalities 32 or higher and outperforming previously reported dev set performance with CNN-based embeddings~\cite{kamper2016deep} for all dimensionalities $\ge 16$.

\vspace{-.1in}
\subsection{Effect of training vocabulary}
\label{ssec:vocab}

We might expect the learned embeddings to be more accurate for words that are seen in training than for ones that are not.  Fig.~\ref{fig:varydim_varymin} measures this effect by showing performance as a function of the number of occurrences of the dev words in the training set.  Indeed, both model types are much more successful for in-vocabulary words, and their performance improves the higher the training frequency of the words. However, performance increases more quickly for the Siamese network than for the classifier as training frequency increases.  This may be due to the fact that, if a word type occurs at least $k$ times in the classifier training set, then it occurs at least $2 \times {k \choose 2}$ times in the Siamese paired training data.

\vspace{-.1in}
\subsection{Visualization of embeddings}
\label{ssec:qual}
 
In order to gain a better qualitative understanding of the differences between clasiffier and Siamese-based embeddings, and of the learned embedding space more generally, we plot a two-dimensional visualization of some of our learned embeddings via t-SNE~\cite{maaten2008visualizing} in Fig.~\ref{fig:tsne}.
For both classifier and Siamese embeddings, there is a marked difference in the quality of clusters formed by embeddings of words that were previously seen vs.~previously unseen in training.  However, the Siamese network embeddings appear to have better relative distances between word clusters with similar and dissimilar pronunciations. For example, the word \texttt{programs} appears equidistant from \texttt{problems} and \texttt{problem} in the classifier-based embedding space, but in the Siamese embedding space \texttt{problems} falls between \texttt{problem} and \texttt{programs}.  Similarly, the cluster for \texttt{democracy} shifts with respect to \texttt{actually} and \texttt{especially} to better respect differences in pronunciation.  More study of learned embeddings, using more data and word types, is needed to confirm such patterns in general. Improvements in unseen word embeddings from the classifier embedding space to the Siamese embedding space (such as for \texttt{democracy}, \texttt{morning}, and \texttt{basketball}) are a likely result of optimizing the model for relative distances between words.

\vspace{-.08in}
\section{Conclusion}
\label{sec:conc}
\vspace{-.05in}

Our main finding is that RNN-based acoustic word embeddings outperform prior approaches, as measured via a word discrimination task related to query-by-example search.  Our best results are obtained with deep LSTM RNNs with a combination of several stacked layers and several fully connected layers, optimized with a contrastive Siamese loss.  Siamese networks have the benefit that, for any given training data set, they are effectively trained on a much larger set, in the sense that they measure a loss and gradient for every possible pair of data points.  Our experiments suggest that the models could still be improved with additional layers.  In addition, we have found that, for the purposes of acoustic word embeddings, fully connected layers are very important and have a more significant effect per layer than stacked layers, particularly when trained with the cross entropy loss function.

These experiments represent an initial exploration of sequential neural models for acoustic word embeddings.  There are a number of directions for further work.  For example, while our analyses suggest that Siamese networks are better than classifier-based models at embedding previously unseen words, our best embeddings are still much poorer for unseen words.  Improvements in this direction may come from larger training sets, or may require new models that better model the shared structure between words.  Other directions for future work include additional forms of supervision and training, as well as application to downstream tasks.

\bibliographystyle{IEEEbib}
\bibliography{refs2}

\end{document}